\documentclass[10pt,twocolumn,letterpaper]{article}

\usepackage{iccv}
\usepackage{times}
\usepackage{epsfig}
\usepackage{graphicx}
\usepackage{amsmath}
\usepackage{amssymb}

\usepackage{multirow}
\usepackage{pifont}
\usepackage{algorithm}
\usepackage{algorithmicx}
\usepackage{algpseudocode}
\usepackage{caption}
\usepackage{subcaption}
\usepackage{setspace}

\usepackage[pagebackref=true,breaklinks=true,letterpaper=true,colorlinks,bookmarks=false]{hyperref} 

\iccvfinalcopy 

\def\iccvPaperID{1345} 

\ificcvfinal\fi

\graphicspath{{figures/}}

\begin{document}\sloppy

  \title{Bridging the Gap: Multi-Level Cross-Modality Joint Alignment for Visible-Infrared Person Re-Identification}

  \author{
    Tengfei Liang$^1$, Yi Jin$^1$, Wu Liu$^2$, Tao Wang$^1$, Songhe Feng$^1$, Yidong Li$^1$ \\
    $^1$School of Computer and Information Technology, Beijing Jiaotong University, Beijing, China \\
    $^2$JD Explore Academy, Beijing, China \\
    {\tt\small \{tengfei.liang,yjin\}@bjtu.edu.cn,liuwu@live.cn,\{twang,shfeng,ydli\}@bjtu.edu.cn}
    \vspace{-5pt}
  }


  \maketitle
  \ificcvfinal\thispagestyle{empty}\fi

  \begin{abstract}
    Visible-Infrared person Re-IDentification (VI-ReID) is a challenging cross-modality image retrieval task that aims to match pedestrians' images across visible and infrared cameras. 
    To solve the modality gap, existing mainstream methods adopt a learning paradigm converting the image retrieval task into an image classification task with cross-entropy loss and auxiliary metric learning losses. 
    These losses follow the strategy of adjusting the distribution of extracted embeddings to reduce the intra-class distance and increase the inter-class distance. 
    However, such objectives do not precisely correspond to the final test setting of the retrieval task, resulting in a new gap at the optimization level. 
    By rethinking these keys of VI-ReID, we propose a simple and effective method, the Multi-level Cross-modality Joint Alignment (MCJA), bridging both modality and objective-level gap. 
    For the former, we design the Modality Alignment Augmentation, which consists of three novel strategies, the weighted grayscale, cross-channel cutmix, and spectrum jitter augmentation, effectively reducing modality discrepancy in the image space. 
    For the latter, we introduce a new Cross-Modality Retrieval loss. 
    It is the first work to constrain from the perspective of the ranking list, aligning with the goal of the testing stage. 
    Moreover, based on the global feature only, our method exhibits good performance and can serve as a strong baseline method for the VI-ReID community.
  \end{abstract}

  \section{Introduction}  \label{sec_introduction}

  \begin{figure}[t]
    \centering
    \includegraphics[width=\linewidth]{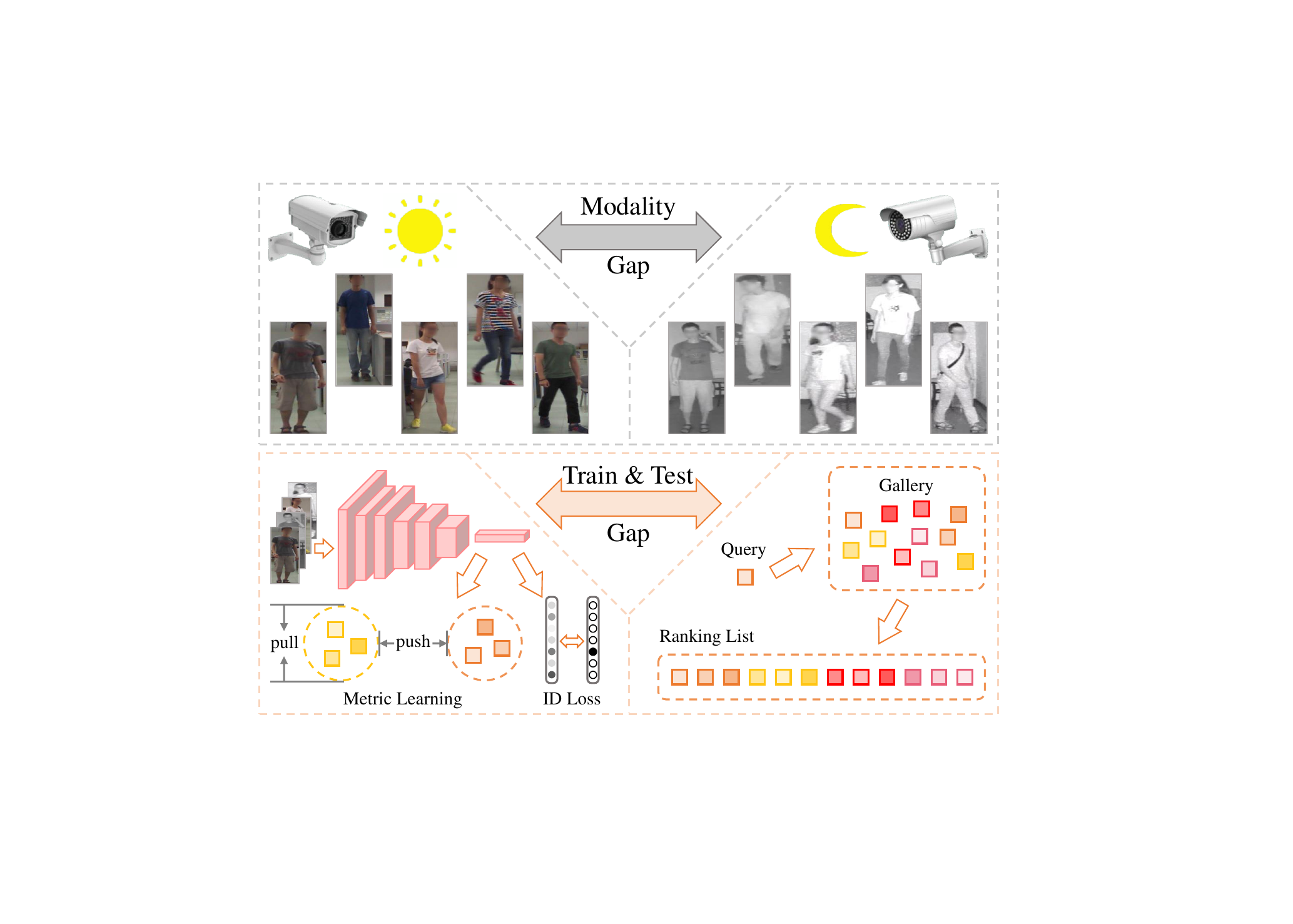}
    \caption{
      Illustration about the gap in VI-ReID task, including the modality gap between visible and infrared images, and the gap between training and testing objectives.
    }
    \label{fig_task_gap}
    \vspace{-10pt}
  \end{figure}

  Person Re-IDentification (ReID)~\cite{arXiv_ReID_Survey} aims at retrieving images of the specific person from a collection of images captured by multiple non-overlapping cameras.
  It has significant value and a wide range of practical applications in the field of security surveillance.
  Despite many years of study and development in this research field~\cite{NeuroComputing19_ReID_Review,TPAMI21_AGW}, existing high-performance ReID methods~\cite{ECCV18_ReID_PCB,ICCV19_ReID_ABD-Net, ICCV19_ReID_Auto-ReID,AAAI20_Viewpoint_Aware_Loss} mainly focus on single visible modality under RGB cameras.
  This restricts their usage to daylight hours with good lighting conditions only.
  To overcome this limitation and achieve continuous 24-hour intelligent surveillance, the recently proposed cross-modality Visible-Infrared person Re-IDentification (VI-ReID) task~\cite{ICCV17_SYSU-MM01,IJCV20_SYSU-MM01} seeks to expand the application scenarios of ReID by matching person images across visible and infrared modalities.
  It is a more challenging task due to the substantial heterogeneous difference between these two modalities.

  In the VI-ReID task, the query and gallery sets consist of images from different modalities, respectively, and it can be regarded as a cross-modality image retrieval task.
  Thus, addressing the modality gap stemming from the different wavelengths of visible and infrared light is a critical challenge in this task.
  Existing methods can be broadly categorized into two groups.
  The first group adopts the generative strategy, transforming one modality into another to reduce the modality gap~\cite{IJCAI18_cmGAN,CVPR19_D2RL,ICCV19_AlignGAN,AAAI20_JSIA} or compensating the information of another modality for the image to obtain a unified multimodal representation~\cite{CVPR20_cm-SSFT,CVPR22_FMCNet}.
  However, due to the lack of a reliable mapping relationship for modality transfer, most of these methods are ineffective, and their adversarial learning training is usually unstable.
  The second group mainly focuses on the design of the model structure and metric learning loss, intending to mine the shared features of pedestrians across modalities.
  In recent years, many such methods~\cite{IJCAI18_BDTR,AAAI19_D-HSME,ECCV20_DDAG,TPAMI21_AGW,CVPR21_MPANet,ICCV21_MCLNet,ICCV21_CM-NAS} have been proposed by researchers, with a commonly used framework paradigm converting this image retrieval task into an image classification task in the training stage. 
  The cross-entropy loss (or called ID loss~\cite{CVPR19_ReID_BoT}) is adopted to implicitly ensure the intra-class consistency of extracted features, and auxiliary metric learning loss is added to regulate the distribution of embeddings in the feature space.
  The core idea of these losses is to minimize intra-class distances and maximize inter-class distances~\cite{TPAMI21_AGW}.
  Although this framework effectively drives performance improvement, the task conversion simultaneously causes inconsistency between the objectives of the training and testing stages, introducing a new gap for the VI-ReID task.
  Besides, the optimization objectives of these methods often contain too many sub losses~\cite{CVPR19_D2RL,CVPR20_cm-SSFT,CVPR21_FMI,ICCV21_MCLNet,AAAI19_D-HSME,AAAI21_CICL+IAMA}, which brings difficulties for the tuning on new datasets.
  Recently, to further improve performance, some methods put more weight on local features. 
  Nevertheless, the fixed or learned partitioning strategies~\cite{MM21_MMN, CVPR21_MPANet, CVPR22_MAUM} increase the computational cost, which hinders their practical application.
 
  Based on the above observation, we find that the gap in this task mainly includes two aspects: the modality gap between visible and infrared images, and the gap between training and testing objectives.
  As illustrated in Figure~\ref{fig_task_gap}, the former comes from the setting of the VI-ReID task, while the latter comes from the paradigm of mainstream methods.
  Aiming to bridge the gap, we propose a simple and effective method, the Multi-level Cross-modality Joint Alignment (MCJA).
  Specifically, for the modality gap, we propose the Modality Alignment Augmentation (MAA) to align the image space of different modalities.
  Here, we design three kinds of novel data augmentation strategies, including the weighted grayscale, cross-channel cutmix, and spectrum jitter, to alleviate the discrepancy between visible and infrared images from the data space.
  For the gap between training and testing, we propose the Cross-Modality Retrieval (CMR) loss.
  Based on the differentiable ranking~\cite{ICML20_Differentiable_Sort_Rank} and Spearman's footrule distance~\cite{WWW10_Rank_Distance}, it directly simulates the retrieval process between query and gallery during training with ranking-aware objective to align with the testing.
  To our knowledge, it has not been explored in the previous literature, and our method is the first work to explicitly address the inconsistent goals between training and testing in this field.
  Through the multi-level joint alignment mechanism, our method effectively improves the performance.
  Moreover, it is a powerful global feature-based method without complicated partitioning strategies, which will be more suitable for practical application.

  Our main contributions can be summarized as follows:

  \begin{itemize}
    \item We propose the Multi-level Cross-modality Joint Alignment (MCJA) method, rethinking the key factors of VI-ReID and constructing a simple and effective global feature-based method. Extensive experiments demonstrate the superiority of our MCJA method compared with existing VI-ReID methods.
    \item To reduce the modality-level gap, the Modality Alignment Augmentation (MAA) is designed with three novel strategies, the weighted grayscale, cross-channel cutmix, and spectrum jitter augmentation, alleviating discrepancy between visible and infrared images.
    \item To solve the objective-level gap, the ranking-aware Cross-Modality Retrieval (CMR) loss is proposed to simulate the retrieval process between query and gallery during training, which is the first work to unify the goals of training and testing stages explicitly. 
  \end{itemize}

  \section{Related Work}  \label{sec_related_work}

  \textbf{Single-Modality Person ReID.}
  The single-modality person ReID task aims at matching images of pedestrians across non-overlapping visible cameras. 
  This task involves several challenges, including intra-class variability and inter-class confusion caused by variations in viewpoint, illumination, weather conditions, human pose, occlusion, etc.
  In recent years, researchers have proposed many methods to address these challenges~\cite{NeuroComputing19_ReID_Review,TPAMI21_AGW}. 
  Zhu \etal~\cite{AAAI20_Viewpoint_Aware_Loss} proposed the Viewpoint-Aware Loss with Angular Regularization to jointly model the viewpoint distribution and identity distribution.
  Xiang \etal~\cite{CVPR22_ReID_Illumination_Rethinking} quantitatively analyzed the influence of illumination on the ReID system and explicitly dissected person ReID from the aspect of illumination on the SynPerson dataset.
  To address weather-related challenges, Li \etal~\cite{MM21_ReID_Weather_WePerson} proposed a Weather Person pipeline to synthesize a new ReID dataset with different weather scenes and designed an adaptive sample selection strategy to close these domain gaps.
  In addition, Su \etal~\cite{ICCV17_ReID_Pose-Driven} designed a Pose-driven Deep Convolutional model to address pose deformations, while Yan \etal~\cite{ICCV21_ReID_Occluded_OPReID} introduced a new model to learn single-scale global representations with the single network backbone for the occluded person ReID. 
  However, despite these advances, all these methods focus on images of visible modality and are thus limited in their applicability to scenarios with poor lighting conditions.

  \begin{figure*}[t]
    \centering
    \includegraphics[width=\linewidth]{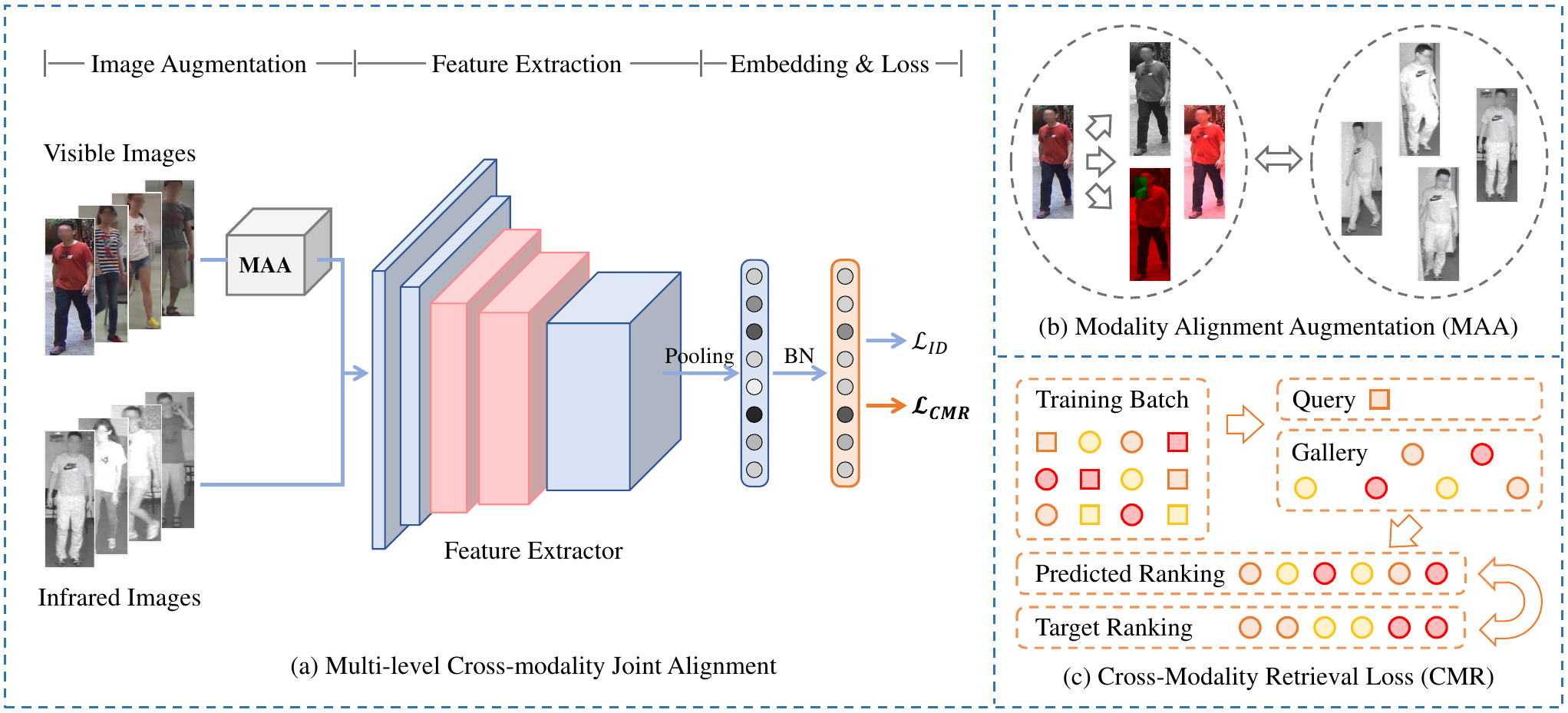}
    \caption{
      The overview of our Multi-level Cross-modality Joint Alignment (MCJA) method. 
      Here, the Modality Alignment Augmentation and Cross-Modality Retrieval Loss are the keys to bridging the modality and objective-level gap for the VI-ReID task.
      In subfig-c, the same color indicates the same ID and the same shape represents the same modality. 
    }
    \label{fig_overall_structure}
  \end{figure*}

  \textbf{Visible-Infrared Person ReID.}
  The visible-infrared person ReID~\cite{ICCV17_SYSU-MM01} aims at addressing the limitations of single-modality ReID by matching pedestrians across cameras with different modalities, i.e., visible and infrared. 
  To solve the gap between these modalities, some researchers proposed cross-modality generation or compensation strategies~\cite{CVPR19_D2RL, AAAI20_JSIA, CVPR20_Hi-CMD}.
  The cross-modality Generative Adversarial Network (cmGAN)\cite{IJCAI18_cmGAN} was the first to explore learning cross-modality representations through the generative adversarial training.
  Then Wang \etal~\cite{ICCV19_AlignGAN} introduced the Alignment Generative Adversarial Network (AlignGAN) that leverages both pixel and feature alignment.
  Lu \etal~\cite{CVPR20_cm-SSFT} designed a cm-SSFT algorithm to exploit both modality-shared and modality-specific characteristics and compensated information of two modalities. 
  Zhang \etal~\cite{CVPR22_FMCNet} proposed the Feature-level Modality Compensation Network (FMCNet), generating missing modality-specific information in the feature space to get a unified multi-modal fusion representation.
  Some other researchers focused on the structure of feature extractors and the design of metric learning losses~\cite{IJCAI18_BDTR, AAAI19_D-HSME, AAAI20_X-Modality, CVPR21_NFS, ICCV21_SMCL, ICCV21_CAJL}.
  Ye \etal~\cite{TPAMI21_AGW} proposed the baseline method, the AGW, with the structure of unshared shallow and shared deep network optimized by weighted regularization triplet loss.
  Then Gao \etal~\cite{MM21_MSO} proposed the MSO method that extended AGW by introducing perceptual edge features loss and cross-modality contrastive-center loss to optimize multi-feature space, while Hao \etal~\cite{ICCV21_MCLNet} added the modality confusion learning strategy and marginal center aggregation losses.
  Besides, to achieve higher performance, the local feature-based approach has gained attention in some recent works, using fixed or learnable partitioning strategies\cite{MM21_MMN, CVPR21_MPANet, CVPR22_MAUM} or prior knowledge guided by human keypoints~\cite{MM22_KMDL}.
  Different from existing methods, we address the gap of VI-ReID via joint alignment in multiple levels on both image modality and optimization objective, and contribute an effective global features-based method for this task.

  \section{The Proposed Method}  \label{sec_method}
  In this section, we present a detailed introduction of our proposed Multi-level Cross-modality Joint Alignment (MCJA) method.
  We begin by providing an overview of our framework in Section~\ref{sec_method_overview}.
  Subsequently, we introduce the designed strategies of Modality Alignment Augmentation for bridging the modality gap in Section.~\ref{sec_method_maa} and the Cross-Modality Retrieval Loss for aligning the optimization objectives of training and testing in Section.~\ref{sec_method_cmr_loss}.
  Lastly, we show the overall objective in Section.~\ref{sec_method_overall_objective}.

  \subsection{Overview}  \label{sec_method_overview}
  In VI-ReID, the data comprises images from visible and infrared modalities.
  Let $X^{m}=\{x^{m}|x^{m} \in \mathbb{R}^{c\times h \times w}, m \in \{vis, ir\}\}$ represent the visible and infrared image sets, where $c$, $h$, and $w$ denote the image's channel, height, and width, respectively.
  During training, a mini-batch is loaded by randomly sampling $P$ pedestrians, with each person represented by $K$ images.
  Therefore, the total number of images in a training batch, denoted by $B$, is equal to $P \times K$.
  
  As depicted in Figure~\ref{fig_overall_structure}a, the first step involves processing the input images with image augmentation.
  In this stage, we design the Modality Alignment Augmentation (MAA, $F_{MAA}$) for visible images to reduce the gap with infrared images, aligning modalities in the original image space.
  Then infrared images $X^{ir}$ and augmented visible images $\hat{X}^{vis}$ are put together to form the input set $\hat{X}^m$:
  \begin{equation}
    \hat{X}^{vis} = F_{MAA}(X^{vis}),
    \quad
    \hat{X}^m = \{\hat{X}^{vis}, X^{ir}\}.
  \end{equation}
  In the stage of feature extraction, the ResNet~\cite{CVPR16_ResNet} model is adopted as the backbone network (denoted as $F_{B}$).
  We follow the existing methods~\cite{TPAMI21_AGW, ICCV21_MCLNet, MM21_MSO, CVPR18_NonLocal} and incorporate non-local blocks into the network.
  The feature maps $f^m$ of each image $\hat{x}^m$ can be extracted as follows:
  \begin{equation}
    f^m = F_{B}(\hat{x}^m),
    \quad
    \hat{x}^m \in \hat{X}^m.
  \end{equation}
  Subsequently, image embeddings are extracted by performing global pooling and batch normalization operations:
  \begin{equation}
    e^m = BatchNorm(Pooling(f^m)),
  \end{equation}
  During training, the whole model is optimized end-to-end by ID loss~\cite{CVPR19_ReID_BoT, TPAMI21_AGW} and our proposed Cross-Modality Retrieval loss.
  During testing, extracted embeddings $e^m$ serve as the final representations for query and gallery images for matching.
  Note that we remove redundant and complex designs and only use global features, aiming to find key factors and construct a simple and effective method.
  
  \subsection{Modality Alignment Augmentation}  \label{sec_method_maa}
  Visible images consist of three channels, namely red, green, and blue, while infrared images are single-channel images that capture the intensity of infrared radiation. 
  To unify the dimensions of input data for a CNN model, a commonly used processing strategy is to simply repeat the single channel of the infrared image three times to obtain three-channel image data. 
  However, from a computational perspective, this leads to significant numerical differences between input images of different modalities. 
  Specifically, the three channels of a visible image are different, while the three channels of an infrared image are identical, resulting in a clear modality gap in the data source.

  To establish a transition between visible and infrared images, we can directly convert the former to grayscale images to obtain a single-channel representation similar to infrared images. 
  Our experiments verify that this strategy can bring a slight performance improvement.
  Nevertheless, this strategy has two drawbacks.
  Firstly, the rich color information from the original three channels of visible images is lost.
  Secondly, the common algorithm for converting to grayscale uses fixed weights to fuse the three channels. 
  Specifically, its calculation formula is as follows:
  \begin{equation}
    x^{gs} = 0.299 \times x^r + 0.587 \times x^g + 0.114 \times x^b, 
  \end{equation}
  which leverages the difference in sensitivity of the human vision to different colors.
  However, from the perspective of the VI-ReID model, the utilization of this prior knowledge is not reasonable.
  Considering that, we design the Modality Alignment Augmentation (MAA), which includes three novel augmentation strategies (Figure~\ref{fig_overall_structure}b). 
  They offer two advantages, as they can generate images similar to the representation of infrared images,  reducing the modality gap, while also fully utilizing the color information of original visible images.
  We introduce each of them as follows.

  \begin{figure}[t]
    \centering
    \begin{subfigure}{\linewidth}
      \centering
      \includegraphics[width=\linewidth]{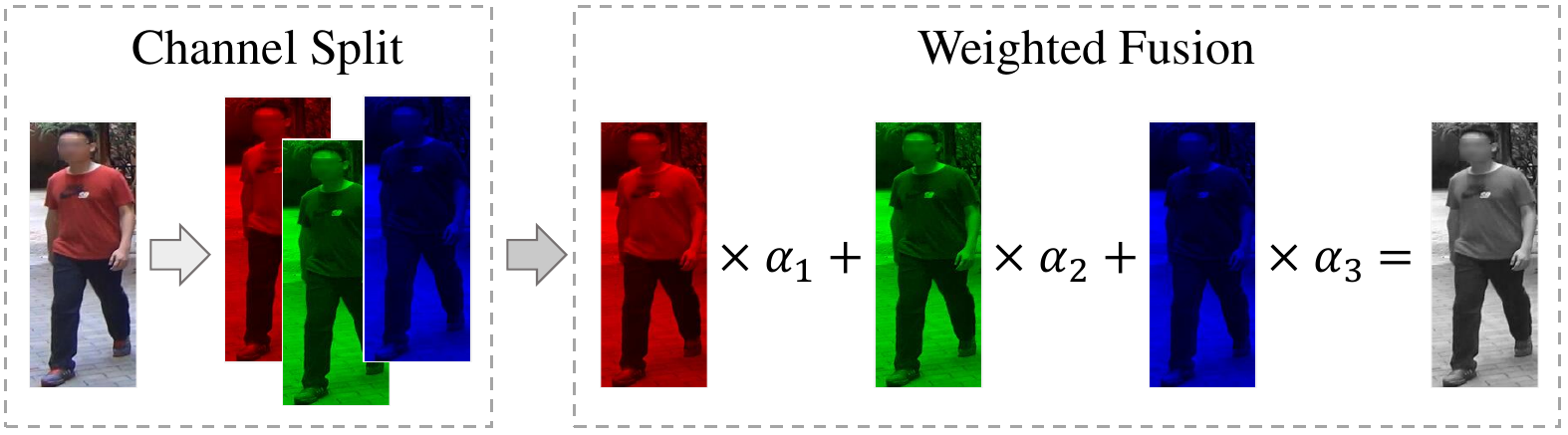}
      \caption{Weighted Grayscale Augmentation}
      \hspace{1pt}
      \label{fig_aug_wg}
    \end{subfigure}
    \centering
    \begin{subfigure}{\linewidth}
      \centering
      \includegraphics[width=\linewidth]{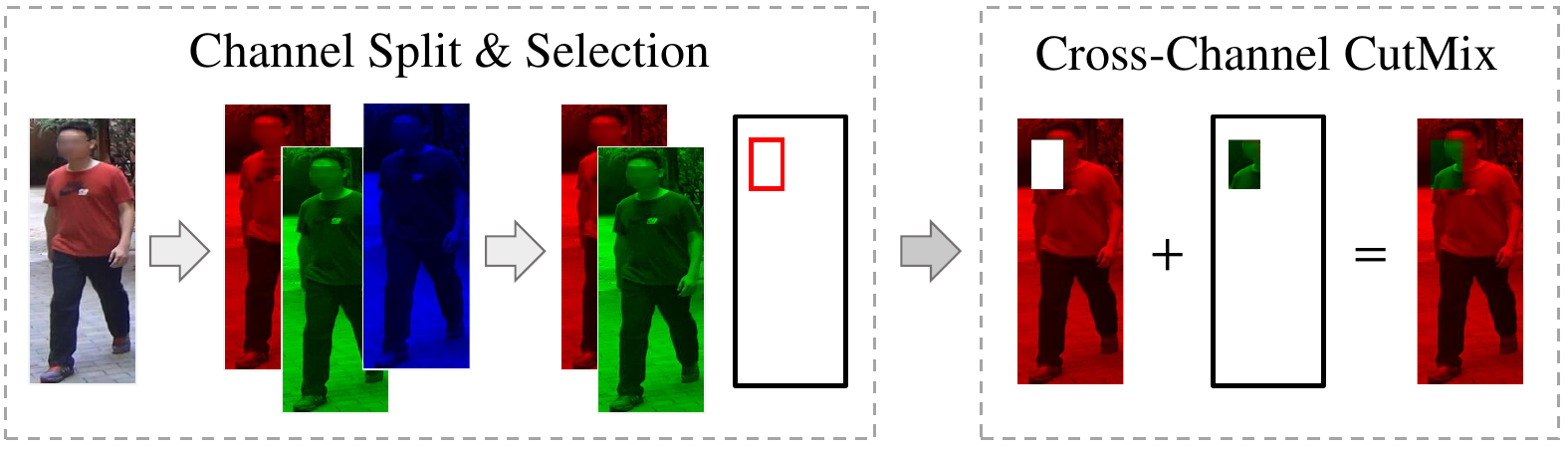}
      \caption{Cross-channel Cutmix Augmentation}
      \hspace{1pt}
      \label{fig_aug_cc}
    \end{subfigure}
    \centering
    \begin{subfigure}{\linewidth}
      \centering
      \includegraphics[width=\linewidth]{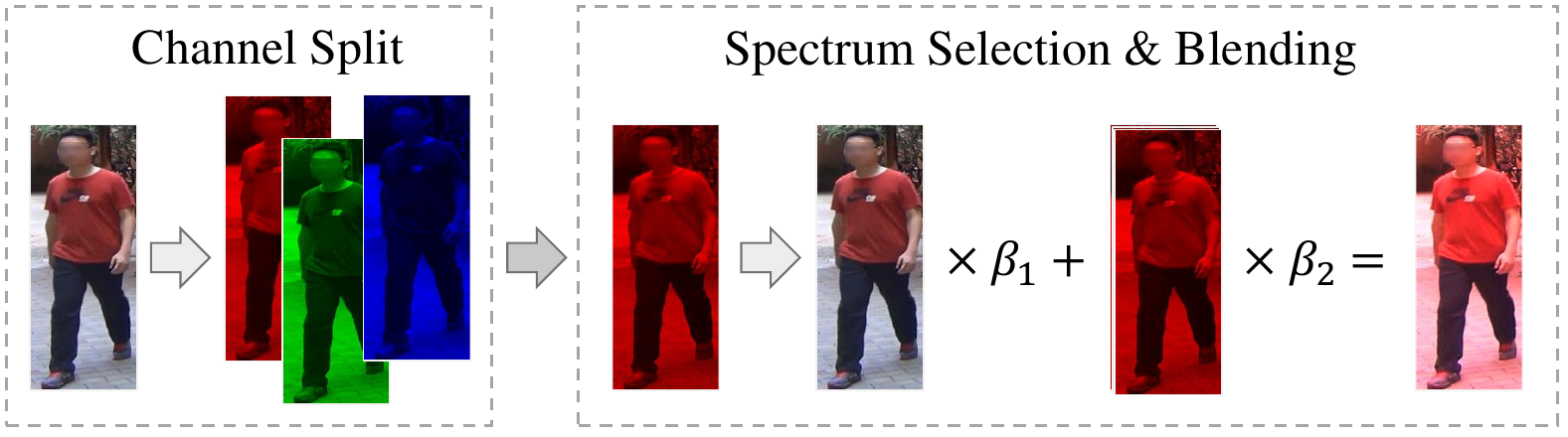}
      \caption{Spectrum Jitter Augmentation}
      \label{fig_aug_sj}
    \end{subfigure}
    \caption{
      The proposed strategies in the Modality Alignment Augmentation. 
      Note that the color of the single-channel image is just for illustration.
    }
    \label{fig_maa}
  \end{figure}

  \textbf{Weighted Grayscale Augmentation.}
  To overcome the limitations of the traditional grayscale algorithm based on prior knowledge, we design a random multi-channel weighted fusion mechanism.
  As shown in Figure~\ref{fig_aug_wg}, a visible image is split into three channels (\textit{i.e.}, $x^r$, $x^g$, $x^b$), and then they are fused using the following formula:
  \begin{equation}
    x^{wg} = \alpha_1 \times x^r + \alpha_2 \times x^g + \alpha_3 \times x^b, 
  \end{equation}
  where the $\alpha_1$, $\alpha_2$ and $\alpha_3$ are randomly generated with values in the range of [0,1], and their sum is 1.
  The $x^{wg}$ denotes the augmented grayscale image.
  It is worth noting that the authors of \cite{ICCV21_CAJL} propose a Channel Exchangeable Augmentation for this task, which randomly selects one channel as the enhanced image. 
  It can be seen as a degraded case of our strategy, where the weight $\alpha$ of one channel is 1, and the weights of other channels are 0. 
  Thus, by using the Weighted Grayscale (WG) augmentation, we can generate various types of grayscale images in a more flexible way.

  \textbf{Cross-channel Cutmix Augmentation.}
  The existing CutMix~\cite{ICCV19_CutMix} method is an augmentation technique that combines multiple images with soft labels for optimization. 
  In contrast, our Cross-channel Cutmix (CC) Augmentation aims at fusing information across multiple channels of the single visible image. 
  As illustrated in Figure~\ref{fig_aug_cc}, we select two channels from the separated channels as foreground and background, respectively. 
  Then, they are combined into a single image $x^{cc}$ by cutting a random rectangular patch from the foreground $x^{fg}$ and pasting it into the background $x^{bg}$.
  It can be formulated as follows:
  \begin{equation}
    \begin{aligned}
      x^{bg}, x^{fg} &= RandSelect(x^{r}, x^{g}, x^{b}) \\
      x^{cc} &= CutMix(x^{bg}, x^{fg})
    \end{aligned}
  \end{equation}
  Note that the foreground and background are randomly selected each time.
  In this way, the color information of different channels is effectively combined.

  \textbf{Spectrum Jitter Augmentation.}
  Our Spectrum Jitter (SJ) strategy is inspired by the existing color jitter augmentation. 
  For example, the brightness adjustment in color jitter is implemented by a weighted blending of the original image and an all-black image~\cite{NIPS19_PyTorch}, and the latter is considered a degenerate image at the brightness level. 
  With a different goal, our method aims to provide spectral adjustment for visible images. 
  Considering that the three channels reflect the radiance intensity in a specific wavelength range of visible light, we construct the degenerate image at the spectral level from one single channel. 
  As shown in Figure~\ref{fig_aug_sj}, we first randomly select a channel and repeat it three times to obtain the degenerate spectral image $x^{ds}$. 
  Then we blend it with the original image as follows:
  \begin{equation}
      x^{sj} = \beta_1 \times x^{vis} + \beta_2 \times x^{ds}, 
  \end{equation}
  where the $\beta_1$ denotes the weight factor with a random value in the range of [0,1] and $\beta_2 = 1 - \beta_1$.
  It gives the degenerate spectral image when $\beta_1$ is 0 and gives the original image when $\beta_1$ is 1.
  When taking other values, it provides a transitional representation of spectral enhancement.

  \subsection{Cross-Modality Retrieval Loss}   \label{sec_method_cmr_loss}
  During training, mainstream VI-ReID methods follow the paradigm of~\cite{CVPR19_ReID_BoT, TPAMI21_AGW}. 
  As illustrated in Figure~\ref{fig_task_gap}, they convert this image retrieval task into a classification task and optimize the model using the cross-entropy loss (also called ID loss in the ReID field) and the metric learning loss. 
  The ID loss acts on the global embeddings extracted after the last BN layer.
  It uses learnable weights of the fully connected layer in the classifier to take the dot product with embeddings, implicitly learning the prototype of each class and constraining the cosine similarity of intra-class samples. 
  The metric learning loss is added to adjust extracted embeddings before the last BN layer after pooling, based on the Euclidean distance, to pull the intra-class samples and push away inter-class samples in the feature space. 
  During testing, the setting is defined as retrieving samples with the same ID as the query from the cross-modality gallery set and generating the corresponding ranking list. 
  There exists a gap between the training and testing objectives.


  To align the objectives of the training and testing stages, we propose the Cross-Modality Retrieval (CMR) loss, which simulates cross-modal retrieval settings within a batch during training and optimizes from the perspective of the ranking results (illustrated in Figure~\ref{fig_overall_structure}c).
  We take each sample in the batch as the query $q_i$ in turn, and construct the gallery set $\mathcal{G}$ with samples of another modality. 
  Then we calculate the cosine distance $D_i$ between them:
  \begin{equation}
    \begin{aligned}
      D_i &= [d_1, d_2, d_3, \dots, d_j, \dots, d_n], \\
      d_j ~ &= \Big(1 - \frac{q_i \cdot g_j}{\|q_i\| \cdot \|g_j\|}\Big) / 2, ~~ g_j \in \mathcal{G}, \\
    \end{aligned}
    \label{eq_cmr_qg_dist}
  \end{equation}
  where $n$ represents the number of samples in the gallery set $\mathcal{G}$.
  By employing the differentiable ranking operation~\cite{ICML20_Differentiable_Sort_Rank} (denoted as $\phi(\cdot)$), we generate the ranking list $R_i$ in ascending order based on the distance matrix $D_i$:
  \begin{equation}
    R_i = \phi(D_i) = [r_1, r_2, r_3, \dots, r_j, \dots, r_n],
    \label{eq_cmr_diff_rank}
  \end{equation}
  where $r_j$ denotes the predicted position for the $j$th sample in the gallery set.
  Considering that the goal during the testing stage is to arrange the samples with the same ID at the front and those with different IDs at the back, we define the target ranking list $\hat{R}_i$ inspired by this rule as follows:
  \begin{equation}
    \begin{aligned}
      \hat{R}_i &= [\hat{r}_1, \hat{r}_2, \hat{r}_3, \dots, \hat{r}_j, \dots, \hat{r}_n], \\
      \hat{r}_j ~ &= 
      \begin{cases}
        ~~ 1,  & \text{if }y^{id}_{g_j} = y^{id}_{q_i}, \\
        ~~ n,  & otherwise,
      \end{cases}
    \end{aligned}
    \label{eq_cmr_target_rank}
  \end{equation}
  where $y^{id}_{q_i}$ and $y^{id}_{g_j}$ represent the ID label of query $q_i$ and gallery $g_j$ respectively.
  We take the Spearman's footrule~\cite{WWW10_Rank_Distance} as the measure of the distance between the predicted ranking $R_i$ and target ranking $\hat{R}_i$:
  \begin{equation}
    d^{rank}_i = \frac{1}{n} \sum^n_{j=1} | r_j - \hat{r}_j |,
    \label{eq_cmr_rank_measure}
  \end{equation}
  Our CMR loss is defined as the mean of the rank distances calculated with all query samples in a training batch:
  \begin{equation}
    \mathcal{L}_{CMR} = \frac{1}{B} \sum^B_{i=1} d^{rank}_i.
    \label{eq_cmr}
  \end{equation}
  By directly considering the ranking list, our CMR loss has a distinct advantage over existing metric learning losses and contributes a new perspective to solve the VI-ReID task.

  \begin{table*}[t]
    \centering
    \fontsize{7}{8.3}\selectfont
    \setlength{\tabcolsep}{0.72mm}
    \begin{tabular}{l|c|c c c c c|c c c c c|c c c c c|c c c c c}
        \hline
        \multirow{3}{*}{~~~~~~~~Methods} & \multirow{3}{*}{Venue} & \multicolumn{10}{c}{\textit{All-Search}} & \multicolumn{10}{|c}{\textit{Indoor-Search}} \cr
        \cline{3-22}
        & ~ & \multicolumn{5}{c}{\textit{Single-Shot}} & \multicolumn{5}{|c}{\textit{Multi-Shot}} & \multicolumn{5}{|c}{\textit{Single-Shot}} & \multicolumn{5}{|c}{\textit{Multi-Shot}} \cr
        & ~ & R1 & R10 & R20 & mAP & mINP & R1 & R10 & R20 & mAP & mINP & R1 & R10 & R20 & mAP & mINP & R1 & R10 & R20 & mAP & mINP \cr
        \hline
        cmGAN\cite{IJCAI18_cmGAN} & IJCAI 18 & 26.97 & 67.51 & 80.56 & 27.80 &   -   & 31.49 & 72.74 & 85.01 & 22.27 &   -   & 31.63 & 77.23 & 89.18 & 42.19 &   -   & 37.00 & 80.94 & 92.11 & 32.76 &   -   \cr
        D$^2$RL\cite{CVPR19_D2RL} & CVPR 19 & 28.90 & 70.60 & 82.40 & 29.20 &   -   &   -   &   -   &   -   &   -   &   -   &   -   &   -   &   -   &   -   &   -   &   -   &   -   &   -   &   -   &   -   \cr
        AlignGAN\cite{ICCV19_AlignGAN} & ICCV 19 & 42.40 & 85.00 & 93.70 & 40.70 &   -   & 51.50 & 89.40 & 95.70 & 33.90 &   -   & 45.90 & 87.60 & 94.40 & 54.30 &   -   & 57.10 & 92.70 & 97.40 & 45.30 &   -   \cr
        JSIA\cite{AAAI20_JSIA} & AAAI 20 & 38.10 & 80.70 & 89.90 & 36.90 &   -   & 45.10 & 85.70 & 93.80 & 29.50 &   -   & 43.80 & 86.20 & 94.20 & 52.90 &   -   & 52.70 & 91.10 & 96.40 & 42.70 &   -   \cr
        Hi-CMD\cite{CVPR20_Hi-CMD} & CVPR 20 & 34.94 & 77.58 &   -   & 35.94 &   -   &   -   &   -   &   -   &   -   &   -   &   -   &   -   &   -   &   -   &   -   &   -   &   -   &   -   &   -   &   -   \cr
        cm-SSFT$^{SQ}$\cite{CVPR20_cm-SSFT} & CVPR 20 & 47.70 &   -   &   -   & 54.10 &   -   & 57.40 &   -   &   -   & 59.10 &   -   &   -   &   -   &   -   &   -   &   -   &   -   &   -   &   -   &   -   &   -   \cr
        cm-SSFT$^{AQ}$\cite{CVPR20_cm-SSFT} & CVPR 20 & 61.60 & 89.20 & 93.90 & 63.20 &   -   & 63.40 & 91.20 & 95.70 & 62.00 &   -   & 70.50 & 94.90 & 97.70 & 72.60 &   -   & 73.00 & 96.30 & 99.10 & 72.40 &   -   \cr
        FMCNet\cite{CVPR22_FMCNet} & CVPR 22 & 66.34 &   -   &   -   & 62.51 &   -   & 73.44 &   -   &   -   & 56.06 &   -   & 68.15 &   -   &   -   & 74.09 &   -   & 78.86 &   -   &   -   & 63.82 &   -   \cr
        \hline
        Zero-Padding\cite{ICCV17_SYSU-MM01} & ICCV 17 & 14.80 & 54.12 & 71.33 & 15.95 &   -   & 19.13 & 61.40 & 78.41 & 10.89 &   -   & 20.58 & 68.38 & 85.79 & 26.92 &   -   & 24.43 & 75.86 & 91.32 & 18.64 &   -   \cr
        BDTR\cite{IJCAI18_BDTR} & IJCAI 18 & 17.01 & 55.43 & 71.96 & 19.66 &   -   &   -   &   -   &   -   &   -   &   -   &   -   &   -   &   -   &   -   &   -   &   -   &   -   &   -   &   -   &   -   \cr
        D-HSME\cite{AAAI19_D-HSME} & AAAI 19 & 20.68 & 62.74 & 77.95 & 23.12 &   -   &   -   &   -   &   -   &   -   &   -   &   -   &   -   &   -   &   -   &   -   &   -   &   -   &   -   &   -   &   -   \cr
        DFE\cite{MM19_DFE} & MM 19 & 48.71 & 88.86 & 95.27 & 48.59 &   -   & 54.63 & 91.62 & 96.83 & 42.14 &   -   & 52.25 & 89.86 & 95.85 & 59.68 &   -   & 59.62 & 94.45 & 98.07 & 50.60 &   -   \cr
        MAC\cite{MM19_MAC} & MM 19 & 33.26 & 79.04 & 90.09 & 36.22 &   -   &   -   &   -   &   -   &   -   &   -   &   -   &   -   &   -   &   -   &   -   &   -   &   -   &   -   &   -   &   -   \cr
        \textit{X}-Modality\cite{AAAI20_X-Modality} & AAAI 20 & 49.92 & 89.79 & 95.96 & 50.73 &   -   &   -   &   -   &   -   &   -   &   -   &   -   &   -   &   -   &   -   &   -   &   -   &   -   &   -   &   -   &   -   \cr
        SIM\cite{IJCAI20_SIM} & IJCAI 20 &   -   &   -   &   -   &   -   &   -   & 56.93 &   -   &   -   & 60.88 &   -   &   -   &   -   &   -   &   -   &   -   &   -   &   -   &   -   &   -   &   -   \cr
        DDAG\cite{ECCV20_DDAG} & ECCV 20 & 54.75 & 90.39 & 95.81 & 53.02 & 39.62 &   -   &   -   &   -   &   -   &   -   & 61.02 & 94.06 & 98.41 & 67.98 & 62.61 &   -   &   -   &   -   &   -   &   -   \cr
        CMM+CML\cite{MM20_CMM+CML} & MM 20 & 51.80 & 92.72 & 97.71 & 51.21 &   -   & 56.27 & 94.08 & 98.12 & 43.39 &   -   & 54.98 & 94.38 & 99.41 & 63.70 &   -   & 60.42 & 96.88 & 99.50 & 53.52 &   -   \cr
        CoAL\cite{MM20_CoAL} & MM 20 & 57.22 & 92.29 & 97.57 & 57.20 &   -   &   -   &   -   &   -   &   -   &   -   & 63.86 & 95.41 & 98.79 & 70.84 &   -   &   -   &   -   &   -   &   -   &   -   \cr
        DG-VAE\cite{MM20_DG-VAE} & MM 20 & 59.49 & 93.77 &   -   & 58.46 &   -   &   -   &   -   &   -   &   -   &   -   &   -   &   -   &   -   &   -   &   -   &   -   &   -   &   -   &   -   &   -   \cr
        AGW\cite{TPAMI21_AGW} & TPAMI 21 & 47.50 & 84.39 & 92.14 & 47.65 & 35.30 &   -   &   -   &   -   &   -   &   -   & 54.17 & 91.14 & 95.98 & 62.97 & 59.23 &   -   &   -   &   -   &   -   &   -   \cr
        CICL+IAMA\cite{AAAI21_CICL+IAMA} & AAAI 21 & 57.20 & 94.30 & 98.40 & 59.30 &   -   & 60.70 & 95.20 & 98.60 & 52.60 &   -   & 66.60 & 98.80 & 99.70 & 74.70 &   -   & 73.80 & 99.40 & 99.90 & 68.30 &   -   \cr
        NFS\cite{CVPR21_NFS} & CVPR 21 & 56.91 & 91.34 & 96.52 & 55.45 &   -   & 63.51 & 94.42 & 97.81 & 48.56 &   -   & 62.79 & 96.53 & 99.07 & 69.79 &   -   & 70.03 & 97.70 & 99.51 & 61.45 &   -   \cr
        FMI\cite{CVPR21_FMI} & CVPR 21 & 60.02 & 94.18 & 98.14 & 58.80 &   -   &   -   &   -   &   -   &   -   &   -   & 66.05 & 96.59 & 99.38 & 72.98 &   -   &   -   &   -   &   -   &   -   &   -   \cr
        MPANet\cite{CVPR21_MPANet} & CVPR 21 & 70.58 & 96.21 & 98.80 & 68.24 &   -   & 75.58 & 97.91 & 99.43 & 62.91 &   -   & 76.74 & 98.21 & 99.57 & 80.95 &   -   & 84.22 & 99.66 & 99.96 & 75.11 &   -   \cr
        MCLNet\cite{ICCV21_MCLNet} & ICCV 21 & 65.40 & 93.33 & 97.14 & 61.98 & 47.39 &   -   &   -   &   -   &   -   &   -   & 72.56 & 96.98 & 99.20 & 76.58 & 72.10 &   -   &   -   &   -   &   -   &   -   \cr
        SMCL\cite{ICCV21_SMCL} & ICCV 21 & 67.39 & 92.87 & 96.76 & 61.78 &   -   & 72.15 & 90.66 & 94.32 & 54.93 &   -   & 68.84 & 96.55 & 98.77 & 75.56 &   -   & 79.57 & 95.33 & 98.00 & 66.57 &   -   \cr
        LbA\cite{ICCV21_LbA} & ICCV 21 & 55.41 &   -   &   -   & 54.14 &   -   &   -   &   -   &   -   &   -   &   -   & 58.46 &   -   &   -   & 66.33 &   -   &   -   &   -   &   -   &   -   &   -   \cr
        CM-NAS\cite{ICCV21_CM-NAS} & ICCV 21 & 61.99 & 92.87 & 97.25 & 60.02 &   -   & 68.68 & 94.92 & 98.36 & 53.45 &   -   & 67.01 & 97.02 & 99.32 & 72.95 &   -   & 76.48 & 98.68 & 99.91 & 65.11 &   -   \cr
        CAJL\cite{ICCV21_CAJL} & ICCV 21 & 69.88 & 95.71 & 98.46 & 66.89 & 53.61 &   -   &   -   &   -   &   -   &   -   & 76.26 & 97.88 & 99.49 & 80.37 & 76.79 &   -   &   -   &   -   &   -   &   -   \cr
        MSA\cite{IJCAI21_MSA} & IJCAI 21 & 63.13 &   -   &   -   & 59.22 &   -   &   -   &   -   &   -   &   -   &   -   & 67.18 &   -   &   -   & 72.74 &   -   &   -   &   -   &   -   &   -   &   -   \cr
        MSO\cite{MM21_MSO} & MM 21 & 58.70 & 92.06 & 97.20 & 56.42 &   -   & 65.85 & 94.37 & 98.25 & 49.56 &   -   & 63.09 & 96.61 & 99.05 & 70.31 &   -   & 72.06 & 97.77 & 99.67 & 61.69 &   -   \cr
        MMN\cite{MM21_MMN} & MM 21 & 70.60 & 96.20 & 99.00 & 66.90 &   -   &   -   &   -   &   -   &   -   &   -   & 76.20 & 97.20 & 99.30 & 79.60 &   -   &   -   &   -   &   -   &   -   &   -   \cr
        MID\cite{AAAI22_MID} & AAAI 22 & 60.27 & 92.90 &   -   & 59.40 &   -   &   -   &   -   &   -   &   -   &   -   & 64.86 & 96.12 &   -   & 70.12 &   -   &   -   &   -   &   -   &   -   &   -   \cr
        MAUM$^G$\cite{CVPR22_MAUM} & CVPR 22 & 61.59 &   -   &   -   & 59.96 &   -   &   -   &   -   &   -   &   -   &   -   & 67.07 &   -   &   -   & 73.58 &   -   &   -   &   -   &   -   &   -   &   -   \cr
        MAUM$^P$\cite{CVPR22_MAUM} & CVPR 22 & 71.68 &   -   &   -   & 68.79 &   -   &   -   &   -   &   -   &   -   &   -   & 76.97 &   -   &   -   & 81.94 &   -   &   -   &   -   &   -   &   -   &   -   \cr
        DCLNet\cite{MM22_DCLNet} & MM 22 & 70.79 & 95.45 & 98.61 & 65.18 &   -   &   -   &   -   &   -   &   -   &   -   & 73.51 &   -   &   -   & 76.80 &   -   &   -   &   -   &   -   &   -   &   -   \cr
        \hline
        \textbf{MCJA}~~ (Ours) & - & \textbf{74.48} & \textbf{96.99} & \textbf{99.31} & \textbf{71.34} & \textbf{58.55} & \textbf{79.99} & \textbf{98.73} & \textbf{99.80} & \textbf{66.62} & \textbf{23.55} & \textbf{82.79} & \textbf{98.88} & \textbf{99.92} & \textbf{85.26} & \textbf{81.86} & \textbf{88.71} & \textbf{99.80} & \textbf{100.00} & \textbf{80.74} & \textbf{48.30} \cr
        \hline
    \end{tabular}
    \caption{Comparison with existing VI-ReID methods on the SYSU-MM01 dataset. 
    Here, we utilize the horizontal line in the table to segregate different types of methods.
    }
    \label{tab_comparison_sysumm01}
  \end{table*}

  \subsection{Overall Objective}  \label{sec_method_overall_objective}
  During training, the overall objective consists of two sub losses: the common ID loss~\cite{CVPR19_ReID_BoT, TPAMI21_AGW} and our proposed CMR loss as follows:
  \begin{equation}
    \mathcal{L} = \mathcal{L}_{ID} + \mathcal{L}_{CMR}.
  \end{equation}
  As shown in Figure~\ref{fig_overall_structure}a, they all constrain the extracted embeddings of the same position (after the last BN layer).
  Unlike existing methods, our objective function only has two sub losses without hyperparameters, which will be more suitable for applications in practical scenarios.
  Note that the $\mathcal{L}_{ID}$ and $\mathcal{L}_{CMR}$ all adjust the feature distribution according to the cosine similarity, and we also use the cosine distance for matching the extracted embeddings in the test phase to maintain the consistency of training and testing.

  \section{Experiments}  \label{sec_experiment}

  \subsection{Experimental Settings}
  \textbf{Datasets.}
  During the experiment, we evaluate our proposed method on two publicly available datasets, SYSU-MM01~\cite{ICCV17_SYSU-MM01,IJCV20_SYSU-MM01} and RegDB~\cite{Sensors17_RegDB}, which are commonly used for quantitative comparison in the VI-ReID field.

  SYSU-MM01~\cite{ICCV17_SYSU-MM01} is a large-scale and challenging dataset collected for the VI-ReID task.
  It consists of 29,033 visible and 15,712 infrared images of 491 identities from 6 cameras, including 4 visible and 2 infrared cameras.
  The training set contains 34,167 images of 395 identities, including 22,258 visible and 11,909 infrared images.
  The test set contains the rest of the dataset from 96 identities.
  During the test, the query set consists of all the infrared images of the test set.
  As for the gallery set, 1 image (or 10 images) of each identity per visible camera are randomly sampled together under the single-shot (or multi-shot) setting.
  Besides, it has two test modes: the all-search mode with all images and the indoor-search mode with only indoor images.

  RegDB~\cite{Sensors17_RegDB} consists of images of 412 identities collected by the aligned visible and far-infrared (thermal) cameras.
  Each identity is captured with 10 visible and 10 thermal images.
  Following the dataset split of most existing methods, 206 identities are randomly sampled for the training set, and the remaining 206 identities' images constitute the test set.
  During the test, under the Visible to Thermal mode, the query set consists of all the visible images of the test set, and the gallery set consists of the remaining thermal images.
  The composition of query and gallery sets is opposite under the Thermal to Visible mode to evaluate the cross-modal retrieval capability in a different direction.

  \textbf{Evaluation Metrics.}
  We follow the existing mainstream VI-ReID methods and adopt the standard Cumulative Matching Characteristic (CMC) and mean Average Precision (mAP) as evaluation metrics.
  The CMC shows the proportion of correct matches in Rank-$k$ ($k \in \{1, 10, 20\}$).
  Moreover, following~\cite{TPAMI21_AGW}, we include the mean inverse negative penalty (mINP) as an additional metric.

  \textbf{Implementation Details.}
  We adopt ResNet-50~\cite{CVPR16_ResNet} as the feature extraction backbone with pre-trained weights on the ImageNet~\cite{CVPR09_ImageNet} dataset.
  For the basic data augmentation, we resize input images to a fixed size of $384 \times 192$ and apply some common strategies in VI-ReID task, including the random horizontal flip, random cropping with padding, color jitter, and random erasing~\cite{AAAI20_RandomErasing}.
  In a training mini-batch, we randomly sample 8 identities and 16 images per identity (\textit{i.e.}, $P = 8$, $K = 16$).
  We use the Adam \cite{ICLR15_Adam} optimizer with weight decay set to 0.0005. 
  The initial learning rate is set to 0.00035.
  For the SYSU-MM01 dataset, the model is trained for 140 epochs, with the learning rate decayed by 0.1 and 0.01 at 80, 120 epochs.
  For the RegDB dataset, we train the model for 200 epochs with the decay of the learning rate at epoch 170.

  \subsection{Comparisons with State-of-the-Art Methods}
  We present a quantitative comparison of our method with existing VI-ReID methods.
  These methods are roughly classified into two categories based on whether they employ a generative or modality compensation strategy, which are separated in the table for clarity.
  Note that we mainly compare methods that use global features with similar frameworks for fairness, and we also include some representative methods based on local features.
  We directly adopt the experimental results reported from their original papers.

  \textbf{Comparisons on SYSU-MM01 Dataset.}
  From Table~\ref{tab_comparison_sysumm01}, we can observe that our MCJA method achieves outstanding performance with a rank-1 accuracy of 74.48\% and mAP of 71.34\% under the challenging all-search mode with the single-shot setting. 
  It outperforms widely used baseline AGW~\cite{TPAMI21_AGW} and its improved variants (MCLNet~\cite{ICCV21_MCLNet}, CM-NAS~\cite{ICCV21_MCLNet}, and MSO~\cite{MM21_MSO}) by a large margin. 
  Moreover, it still shows superiority over MPANet~\cite{CVPR21_MPANet}, MMN~\cite{MM21_MMN}, and MAUM$^P$~\cite{CVPR22_MAUM} methods, which use fixed or learnable partitioning strategies to extracted local features and increase the consumption of computing resource.
  The simple and effective design of our method shows more significant advantages compared with existing methods.

  \begin{table}[h]
    \centering
    \fontsize{7}{8.8}\selectfont
    \setlength{\tabcolsep}{1.2mm}
    \begin{tabular}{l|c|c c c|c c c}
        \hline
        \multirow{2}{*}{~~~~~Methods} & \multirow{2}{*}{Venue}
        & \multicolumn{3}{c}{\textit{Visible to Thermal}} & \multicolumn{3}{|c}{\textit{Thermal to Visible}} \cr
        & ~ & R1 & mAP & mINP & R1 & mAP & mINP \cr
        \hline
        D$^2$RL\cite{CVPR19_D2RL} & CVPR 19 & 43.40 & 44.10 &   -   &   -   &   -   &   -   \cr
        AlignGAN\cite{ICCV19_AlignGAN} & ICCV 19 & 57.90 & 53.60 &   -   & 56.30 & 53.40 &   -   \cr
        JSIA\cite{AAAI20_JSIA} & AAAI 20 & 48.50 & 49.30 &   -   & 48.10 & 48.90 &   -   \cr
        Hi-CMD\cite{CVPR20_Hi-CMD} & CVPR 20 & 70.93 & 66.04 &   -   &   -   &   -   &   -   \cr
        cm-SSFT$^{SQ}$\cite{CVPR20_cm-SSFT} & CVPR 20 & 65.40 & 65.60 &   -   & 63.80 & 64.20 &   -   \cr
        cm-SSFT$^{AQ}$\cite{CVPR20_cm-SSFT} & CVPR 20 & 72.30 & 72.90 &   -   & 71.00 & 71.70 &   -   \cr
        FMCNet\cite{CVPR22_FMCNet} & CVPR 22 & 89.12 & 84.43 &   -   & 88.38 & 83.86 &   -   \cr
        \hline
        Zero-Padding\cite{ICCV17_SYSU-MM01} & ICCV 17 & 17.75 & 18.90 &   -   &   -   &   -   &   -   \cr
        HCML\cite{AAAI18_HCML} & AAAI 18 & 24.44 & 20.08 &   -   & 21.70 & 22.24 &   -   \cr
        BDTR\cite{IJCAI18_BDTR} & IJCAI 18 & 33.47 & 31.83 &   -   &   -   &   -   &   -   \cr
        D-HSME\cite{AAAI19_D-HSME} & AAAI 19 & 50.85 & 47.00 &   -   & 50.15 & 46.16 &   -   \cr
        DFE\cite{MM19_DFE} & MM 19 & 70.13 & 69.14 &   -   & 67.99 & 66.70 &   -   \cr
        MAC\cite{MM19_MAC} & MM 19 & 36.43 & 37.03 &   -   &   -   &   -   &   -   \cr
        \textit{X}-Modality\cite{AAAI20_X-Modality} & AAAI 20 &   -   &   -   &   -   & 62.21 & 60.18 &   -   \cr
        SIM\cite{IJCAI20_SIM} & IJCAI 20 & 74.47 & 75.29 &   -   & 75.24 & 78.30 &   -   \cr
        DDAG\cite{ECCV20_DDAG} & ECCV 20 & 69.34 & 63.46 &   -   & 68.06 & 61.80 &   -   \cr
        CMM+CML\cite{MM20_CMM+CML} & MM 20 & 59.81 & 60.86 &   -   &   -   &   -   &   -   \cr
        CoAL\cite{MM20_CoAL} & MM 20 & 74.12 & 69.87 &   -   &   -   &   -   &   -   \cr
        DG-VAE\cite{MM20_DG-VAE} & MM 20 & 72.97 & 71.78 &   -   &   -   &   -   &   -   \cr
        AGW\cite{TPAMI21_AGW} & TPAMI 21 & 80.31 & 73.07 & 57.39 & 75.93 & 69.49 & 52.63 \cr
        CICL+IAMA\cite{AAAI21_CICL+IAMA} & AAAI 21 & 78.80 & 69.40 &   -   & 77.90 & 69.40 &   -   \cr
        NFS\cite{CVPR21_NFS} & CVPR 21 & 80.54 & 72.10 &   -   & 77.95 & 69.79 &   -   \cr
        FMI\cite{CVPR21_FMI} & CVPR 21 & 73.20 & 71.60 &   -   & 71.80 & 70.10 &   -   \cr
        MPANet\cite{CVPR21_MPANet} & CVPR 21 & 83.70 & 80.90 &   -   & 82.80 & 80.70 &   -   \cr
        MCLNet\cite{ICCV21_MCLNet} & ICCV 21 & 80.31 & 73.07 & 57.39 & 75.93 & 69.49 & 52.63 \cr
        SMCL\cite{ICCV21_SMCL} & ICCV 21 & 83.93 & 79.83 &   -   & 83.05 & 78.57 &   -   \cr
        LbA\cite{ICCV21_LbA} & ICCV 21 & 74.17 & 67.64 &   -   & 72.43 & 65.46 &   -   \cr
        CM-NAS\cite{ICCV21_CM-NAS} & ICCV 21 & 84.54 & 80.32 &   -   & 82.57 & 78.31 &   -   \cr
        CAJL\cite{ICCV21_CAJL} & ICCV 21 & 85.03 & 79.14 & 65.33 & 84.75 & 77.82 & 61.56 \cr
        MSA\cite{IJCAI21_MSA} & IJCAI 21 & 84.86 & 82.16 &   -   &   -   &   -   &   -   \cr
        MSO\cite{MM21_MSO} & MM 21 & 73.60 & 66.90 &   -   & 74.60 & 67.50 &   -   \cr
        MMN\cite{MM21_MMN} & MM 21 & 91.60 & 84.10 &   -   & 87.50 & 80.50 &   -   \cr
        MID\cite{AAAI22_MID} & AAAI 22 & 87.45 & 84.85 &   -   & 84.29 & 81.41 &   -   \cr
        MAUM$^G$\cite{CVPR22_MAUM} & CVPR 22 & 83.39 & 78.75 &   -   & 81.07 & 78.89 &   -   \cr
        MAUM$^P$\cite{CVPR22_MAUM} & CVPR 22 & 87.87 & 85.09 &   -   & 86.95 & 84.34 &   -   \cr
        DCLNet\cite{MM22_DCLNet} & MM 22 & 81.20 & 74.30 &   -   & 78.00 & 70.60 &   -   \cr
        \hline
        \textbf{MCJA}~~ (Ours) & - & \textbf{91.80} & \textbf{86.08} & \textbf{74.93} & \textbf{88.06} & \textbf{83.06} & \textbf{70.80} \cr
        \hline
    \end{tabular}
    \caption{Comparison with methods on the RegDB dataset.}
    \label{tab_comparison_regdb}
  \end{table}

  \textbf{Comparisons on RegDB Dataset.}
  As shown in Table~\ref{tab_comparison_regdb}, our method performs well in both test modes, achieving a rank-1 accuracy of 91.80\% and mAP of 86.08\% in the Visible to Thermal mode and a rank-1 accuracy of 88.06\% and mAP of 83.06\% in the Thermal to Visible mode.
  Therefore, our MCJA model exhibits robustness across different query settings, and the multi-level joint alignment effectively improves the cross-modality retrieval capability.
  All these experiments demonstrate the effectiveness of our method.

  \subsection{Ablation Study}
  To better understand our method, we perform ablation experiments on the SYSU-MM01, employing the most difficult all-search mode with single-shot setting by default.

  \textbf{Effectiveness of Each Component.}
  As shown in Table~\ref{tab_ablation_base_maa_cmr}, we adopt the AGW model optimized with only ID loss as the baseline.
  Compared with the reported results in its original paper~\cite{TPAMI21_AGW}, we find that it is significantly underestimated and outperforms many current complex methods, achieving 65.66\% rank-1 accuracy and 63.04\% mAP. 
  This further illustrates the necessity of our contributions to rethinking the key factors of the VI-ReID task. 
  Then we add the proposed MAA and CMR loss, respectively, and they all produce a marked enhancement in performance (Index-2\&3 of Table~\ref{tab_ablation_base_maa_cmr}). 
  It is worth noting that the modality alignment and the alignment of the training and testing objectives are orthogonal.
  Consequently, the joint application of MAA and CMR loss brings further performance improvement.

  \begin{table}[t]
    \centering
    \fontsize{8}{10}\selectfont
    \setlength{\tabcolsep}{1.2mm}
    \begin{tabular}{c|c c c|c c c c c}
        \hline
        Index & Baseline & MAA & CMR & R1 & R10 & R20 & mAP & mINP \cr
        \hline
        1 & \ding{51} & \ding{55}  & ~\ding{55} & 65.66 & 94.50 & 98.06 & 63.04 & 49.48 \cr
        2 & \ding{51} & \ding{51}  & ~\ding{55} & 70.67 & 95.10 & 97.98 & 66.82 & 52.76 \cr
        3 & \ding{51} & \ding{55}  & ~\ding{51} & 68.89 & 95.87 & 99.04 & 67.10 & 54.97 \cr
        \hline
        4 & \ding{51} & \ding{51}  & ~\ding{51} & \textbf{74.48} & \textbf{96.99} & \textbf{99.31} & \textbf{71.34} & \textbf{58.55} \cr
        \hline
    \end{tabular}
    \caption{Ablation experiments for designed components.}
    \label{tab_ablation_base_maa_cmr}
  \end{table}

  \begin{table}[t]
    \centering
    \fontsize{8}{10}\selectfont
    \setlength{\tabcolsep}{1.65mm}
    \begin{tabular}{c|l|c c c c c}
        \hline
        Index & ~~~~~~~~~~~Methods & R1 & R10 & R20 & mAP & mINP \cr
        \hline
        1 & Baseline      & 65.66 & 94.50 & 98.06 & 63.04 & 49.48 \cr
        2 & Baseline + GS & 67.06 & 93.94 & 97.68 & 63.44 & 48.90 \cr
        3 & Baseline + CE~\cite{ICCV21_CAJL} & 68.01 & 94.44 & 97.85 & 64.35 & 49.81 \cr
        \hline
        4 & Baseline + WG & 68.68 & 94.71 & 97.93 & 64.96 & 50.66 \cr
        5 & Baseline + CC & 68.96 & 94.50 & 97.84 & 65.19 & 50.83 \cr
        6 & Baseline + SJ & 67.58 & 94.26 & 97.51 & 65.09 & 51.82 \cr
        \hline
        7 & Baseline + MAA & \textbf{70.67} & \textbf{95.10} & \textbf{97.98} & \textbf{66.82} & \textbf{52.76} \cr
        \hline
    \end{tabular}
    \caption{Ablation study for strategies of the proposed Modality Alignment Augmentation.}
    \label{tab_maa_comparison}
  \end{table}

  \textbf{Discussion on Modality Alignment Augmentation.}
  Our MAA is designed to bridge the modality gap in the image space and consists of three augmentation strategies: Weighted Grayscale (WG), Cross-channel Cutmix (CC), and Spectrum Jitter (SJ). 
  Here, we conduct separate experiments for each strategy in Table~\ref{tab_maa_comparison}. 
  Compared with the baseline, our WG, CC, and SJ each bring a +3.02\%, +3.30\%, +1.92\% rank-1 accuracy and a +1.92\%, +2.15\%, +2.05\% mAP improvement, respectively. 
  Besides, our WG outperforms existing GrayScale (GS, Index-2) and Channel Exchangeable (CE, Index-3) augmentation~\cite{ICCV21_CAJL}, which can be considered as special cases of WG.
  The MAA is implemented by randomly selecting one of WG, CC, and SJ each time, and experimental results (Index-7) demonstrate its effectiveness in aligning visible and infrared modalities.

\begin{figure}[t]
    \centering
    \includegraphics[width=\linewidth]{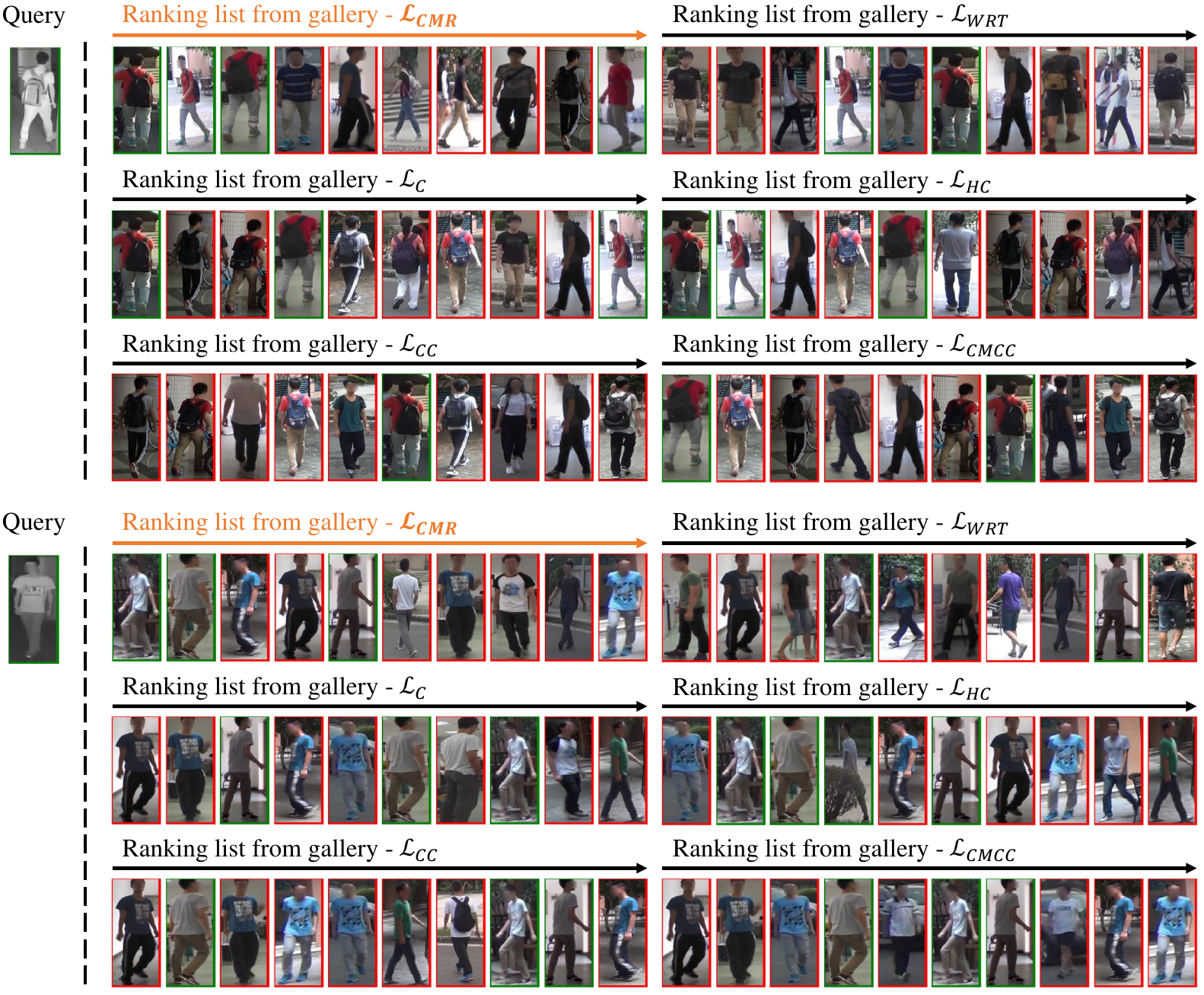}
    \caption{
      The ranking list of retrieval results with different losses. 
      Green indicates the correct match, while red indicates the wrong match.
    }
    \label{fig_ranking_list}
  \end{figure}

  \textbf{Discussion on Cross-Modality Retrieval Loss.}
  Researchers have designed various metric learning losses to adjust the distribution of feature vectors, such as Weighted Regularization Triplet loss ($\mathcal{L}_{WRT}$), Center loss ($\mathcal{L}_{C}$), Hetero-Center (HC) loss ($\mathcal{L}_{HC}$), Center Cluster loss ($\mathcal{L}_{CC}$), and Cross-Modality Contrastive Center loss ($\mathcal{L}_{CMCC}$). 
  These losses basically cover the ideas of mainstream VI-ReID metric learning losses. 
  In Table~\ref{tab_cmr_comparison}, Baseline$^M$ represents the baseline method with our MAA strategy, and we want to observe the effects of different losses on such a strong baseline. 
  Based on quantitative results, we can notice that the performance improvement of $\mathcal{L}_{C}$, $\mathcal{L}_{CC}$, and $\mathcal{L}_{CMCC}$ loss is quite limited, while $\mathcal{L}_{WRT}$ and $\mathcal{L}_{HC}$ perform slightly better. 
  Our ranking-aware $\mathcal{L}_{CMR}$ brings a more substantial performance boost with +3.81\% rank-1 accuracy and +4.52\% mAP.
  From qualitative results in Figure~\ref{fig_ranking_list}, the CMR loss can make samples in the gallery that match the query correctly closer to the front of the list.
  Note that the metric learning losses ignore the overall constraints on ranking results, and only constrain the relative distance between positive and negative pairs in the feature space.
  Our CMR loss overcomes their limitations by providing direct guidance for the ranking list and effectively improves the cross-modality retrieval ability of the method.


  \begin{table}[t]
    \centering
    \fontsize{8}{9}\selectfont
    \setlength{\tabcolsep}{1mm}
    \begin{tabular}{c|l|c c c c c}
        \hline
        Index & ~~~~~~~~~~~Methods & R1 & R10 & R20 & mAP & mINP \cr
        \hline
        1 & Baseline$^{M}$ & 70.67 & 95.10 & 97.98 & 66.82 & 52.76 \cr
        2 & Baseline$^{M}$ + $\mathcal{L}_{WRT}$~\cite{TPAMI21_AGW} & 71.29 & 95.50 & 98.26 & 68.90 & 56.67 \cr
        3 & Baseline$^{M}$ + $\mathcal{L}_{C}$~\cite{ECCV16_Center_Loss} & 70.64 & 95.02 & 97.93 & 66.48 & 52.24 \cr
        4 & Baseline$^{M}$ + $\mathcal{L}_{HC}$~\cite{NeuroComputing20_HC_Loss} & 71.58 & 95.12 & 98.07 & 67.35 & 52.88 \cr
        5 & Baseline$^{M}$ + $\mathcal{L}_{CC}$~\cite{CVPR21_MPANet} & 70.88 & 95.17 & 98.27 & 67.15 & 53.10 \cr
        6 & Baseline$^{M}$ + $\mathcal{L}_{CMCC}$~\cite{MM21_MSO} & 70.96 & 95.28 & 98.25 & 67.24 & 53.27 \cr
        \hline
        7 & Baseline$^{M}$ + $\mathcal{L}_{CMR}$  & \textbf{74.48} & \textbf{96.99} & \textbf{99.31} & \textbf{71.34} & \textbf{58.55} \cr
        \hline
    \end{tabular}
    \caption{Comparison of different losses in VI-ReID task.}
    \label{tab_cmr_comparison}
  \end{table}

  \section{Conclusion}  \label{sec_conclusion}
  In this paper, we propose the Multi-Level Cross-Modality Joint Alignment method for the VI-ReID task. 
  Our method aims to bridge both the modality gap and the gap between training and testing objectives, which are the keys to solving this task. 
  To address the modality-level alignment, we design the Modality Alignment Augmentation, which reduces the modality discrepancy in the original image space. 
  For the objective-level alignment, the Cross-Modality Retrieval loss is proposed to guide the extracted embeddings from the perspective of ranking. 
  Experiments on SYSU-MM01 and RegDB datasets demonstrate the superior performance of our method.
  For future research, we believe that our method can serve as a powerful baseline for the VI-ReID community. 
  We hope this will inspire researchers to propose more effective strategies and advance the development of the field.

  \clearpage
  {\small
  \bibliographystyle{ieee_fullname}
  \begin{spacing}{0.98}
  \bibliography{paper_bib_condensed}
  \end{spacing}
  }

\end{document}